\documentclass[letterpaper]{article} % DO NOT CHANGE THIS
\usepackage[draft]{aaai2026}  % DO NOT CHANGE THIS
\usepackage{times}  % DO NOT CHANGE THIS
\usepackage{helvet}  % DO NOT CHANGE THIS
\usepackage{courier}  % DO NOT CHANGE THIS
\usepackage[hyphens]{url}  % DO NOT CHANGE THIS
\usepackage{graphicx} % DO NOT CHANGE THIS
\usepackage{natbib}  % DO NOT CHANGE THIS AND DO NOT ADD ANY OPTIONS TO IT
\usepackage{caption} % DO NOT CHANGE THIS AND DO NOT ADD ANY OPTIONS TO IT

\usepackage{tabularx, colortbl}
\usepackage[table]{xcolor}
\usepackage{booktabs}
\usepackage{multirow}
\usepackage{xcolor}
\usepackage{arydshln}
\usepackage{multirow}
\usepackage[table]{xcolor}

\usepackage{algorithm}
\usepackage{algorithmic}

\usepackage[most]{tcolorbox}
\usepackage{pifont}
\usepackage{makecell}

\definecolor{promptgreen}{HTML}{006400}
\definecolor{promptbg}{HTML}{F3FCF1}
\usepackage{tabularx}
\usepackage{array}

\definecolor{promptgreen}{HTML}{2E6F40}
\definecolor{promptbg}{HTML}{F7FAF7}

\newtcolorbox{promptbox}[1][]{
    enhanced,
    breakable,
    colback=promptbg,
    colframe=promptgreen,
    colbacktitle=promptgreen,
    coltitle=white,
    title={#1},
    fonttitle=\bfseries,
    fontupper=\small,
    boxrule=0.8pt,
    arc=1mm,
    left=3mm,
    right=3mm,
    top=2mm,
    bottom=2mm,
    before upper={
        \setlength{\parindent}{0pt}
        \setlength{\parskip}{0.35em}
    }
}

\usepackage{newfloat}
\usepackage{listings}
\DeclareCaptionStyle{ruled}{labelfont=normalfont,labelsep=colon,strut=off} % DO NOT CHANGE THIS
\floatstyle{ruled}
\newfloat{listing}{tb}{lst}{}
\floatname{listing}{Listing}
\title{\textsc{LongCrafter}: Towards Diverse Long-Context Understanding via Evidence-Graph-Guided Instruction Synthesis}
\author{
    Chenhao Yuan\textsuperscript{\rm 1}\equalcontrib, 
    Yinhao Xu\textsuperscript{\rm 1}\equalcontrib, 
    Shuwen Xu\textsuperscript{\rm 1}, 
    Xizhi Yang\textsuperscript{\rm 1}, 
    Jiaxiang Liu\textsuperscript{\rm 2}, 
    Chenxi Zhou\textsuperscript{\rm 2}, 
    Shaoping Huang\textsuperscript{\rm 2},
    Haolin Ren\textsuperscript{\rm 1}, 
    Pengfei Cao\textsuperscript{$\dagger$\rm 2}, 
    Jun Zhao\textsuperscript{\rm 2}, 
    Kang Liu\thanks{Corresponding author.}\textsuperscript{\rm 2}
}

\affiliations{
    \textsuperscript{\rm 1}University of Chinese Academy of Sciences, Beijing, China\\
    \textsuperscript{\rm 2}The Key Laboratory of Cognition and Decision Intelligence for Complex Systems,\\
    Institute of Automation, Chinese Academy of Sciences, Beijing, China
    
}

\usepackage{bibentry}
\begin{document}
\makeatletter
\renewcommand\section{\@startsection{section}{1}{\z@}%
  {-2.0ex \@plus -0.5ex \@minus -0.2ex}%
  {1.5ex \@plus 0.3ex \@minus 0.2ex}%
  {\reset@font\Large\bfseries\centering}}
\renewcommand\subsection{\@startsection{subsection}{2}{\z@}%
  {-1.8ex \@plus -0.5ex \@minus -0.2ex}%
  {0.8ex \@plus 0.2ex}%
  {\reset@font\large\bfseries}}
\setcounter{secnumdepth}{2}
\makeatother

\maketitle

\begin{abstract}
Synthesizing long-context supervised fine-tuning (SFT) data offers a scalable way to improve the long-context understanding of large language models (LLMs). However, existing approaches face two major limitations: limited controllability in synthesizing diverse tasks and a lack of explicit faithfulness supervision. This paper proposes \textbf{LongCrafter}, a framework combining a hierarchical task taxonomy with an evidence-grounded pipeline. It broadens coverage with 32 fine-grained long-context task types. To synthesize high-difficulty tasks, LongCrafter further constructs task-aligned contexts and evidence graphs that capture cross-paragraph dependencies, enabling instruction generation at a target difficulty level. To provide explicit faithfulness supervision, it generates responses with cited evidence spans, aligning each reasoning step with its supporting evidence. Data analysis shows that LongCrafter controllably produces diverse high-difficulty tasks and faithful responses. Models trained on this data consistently outperform SFT baselines and match or surpass official post-trained models on LongBench, LongBench v2, and LooGLE. Notably, they achieve particularly substantial gains on high-difficulty tasks.
 These models also remain robust to evidence position, effectively mitigating the ``lost in the middle'' problem.

\end{abstract}

\section{Introduction}

\begin{figure}[!t]
    \centering
    \includegraphics[width=0.9\columnwidth]{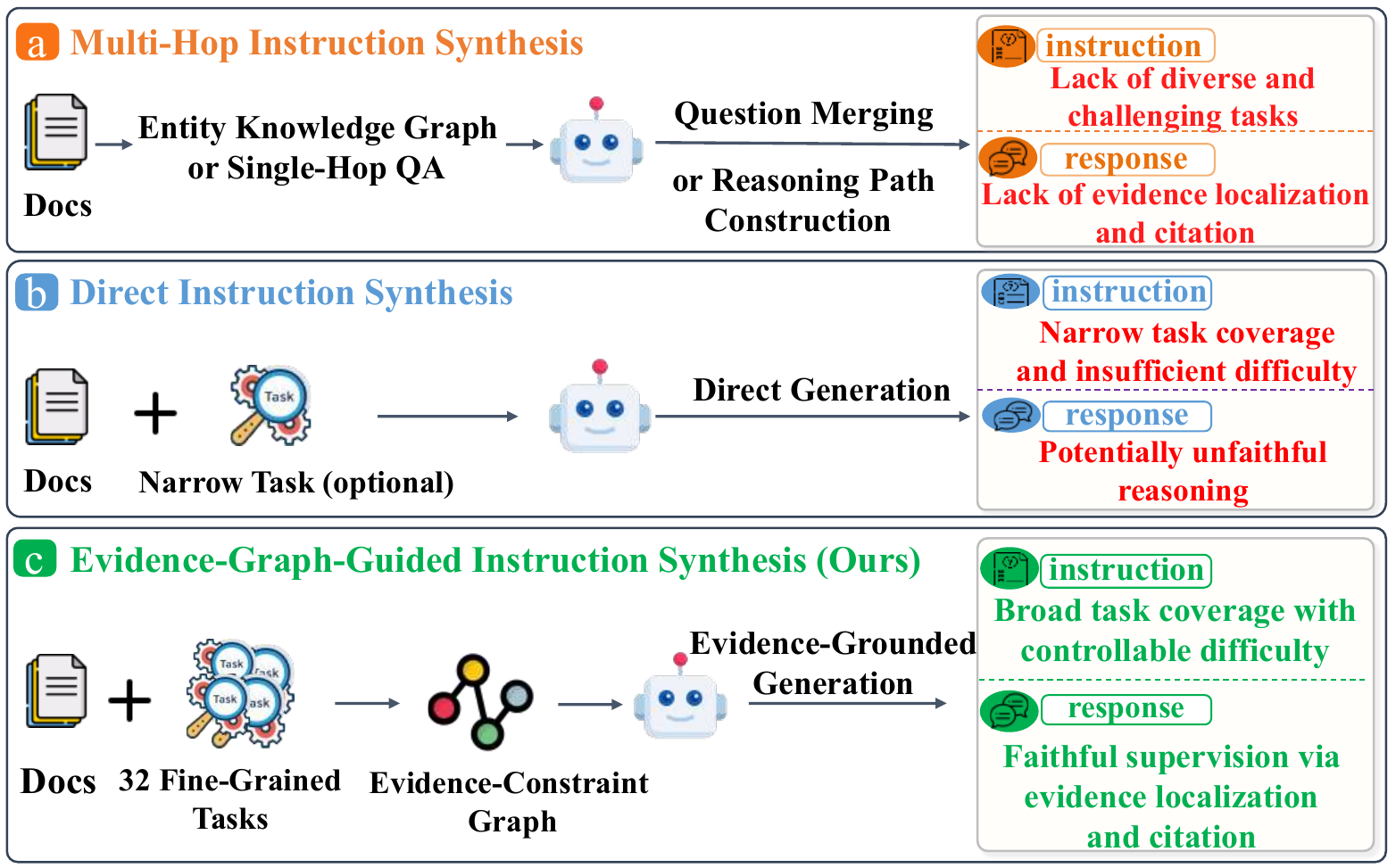}
    \caption{
Comparison of long-context instruction synthesis paradigms.
LongCrafter combines fine-grained task design with evidence-constraint graphs to enable broader task coverage, controllable synthesis of high-difficulty instructions, and explicit faithfulness supervision.
}
    \label{fig:task1}
   
\end{figure}

Long-context understanding has emerged as a critical capability for large language models (LLMs), as real-world applications such as temporal reasoning, aggregation and state tracking require models to recognize and utilize relevant evidence across tens or hundreds of thousands of tokens~\citep{longbench,longbenchv2,loogle,rl}. Supervised fine-tuning (SFT) on synthesized long-context instruction data has emerged as a promising and scalable solution~\citep{longalign,longreward,longmagpie}. Existing synthesis methods generally follow two paradigms, as illustrated in Figure~\ref{fig:task1}. Multi-hop methods derive concept knowledge graphs or single-hop QA pairs from documents and then merge questions or construct reasoning paths~\citep{cgmis,longmit}. In contrast, direct methods generate instruction--response pairs directly from documents, optionally conditioned on a narrow set of predefined tasks~\citep{longalign,longmagpie,longreward}. Despite their different synthesis strategies, both paradigms suffer from two major limitations: \textbf{1) Limited controllability in synthesizing diverse tasks.} 
Current synthesis methods either focus primarily on synthesizing challenging multi-hop questions, resulting in limited task coverage, or directly generate instructions from documents, often producing simple and locally answerable questions~\citep{cgmis,longalign,longmagpie,longreward}. Thus, they lack a unified framework for controllably synthesizing a diverse range of tasks; in particular, they offer limited support for ensuring that tasks beyond multi-hop QA reach their target difficulty. \textbf{2) Lack of explicit faithfulness supervision.} Existing methods either synthesize responses directly from long documents or concept-level knowledge graphs, or provide reasoning with only coarse-grained evidence attribution~\citep{longmagpie,longfaith,cgmis,chain}. Consequently, they offer limited supervision for learning fine-grained evidence localization and performing reasoning strictly grounded in the source text.

To this end, the paper proposes \textbf{LongCrafter}, which frames long-context training data construction as a structured crafting process. To address the aforementioned limitations, LongCrafter introduces a taxonomy of 32 fine-grained task types that broadens the task space considered in prior work, along with a three-stage pipeline built upon this taxonomy: \textit{(1) Long Context Construction}: It matches documents to suitable task types and constructs task-aligned long contexts tailored to their requirements. \textit{(2) Evidence-Constraint Graph Construction}: It organizes task-relevant evidence spans and their cross-paragraph dependencies into a graph, providing the structural constraints needed to meet the reasoning demands of high-difficulty task types. \textit{(3) Evidence-Grounded Instruction-Response Synthesis}: It generates instructions from the evidence graph, ensuring that each instruction reaches the intended difficulty of its target task type. It then generates responses that present step-by-step reasoning, with each step explicitly locating and citing supporting evidence, thereby providing explicit supervision for faithful, evidence-grounded reasoning.

Data analysis shows that LongCrafter supports controllable synthesis across diverse task types and target difficulty levels, while outperforming existing datasets in response faithfulness and other quality metrics. Extensive experiments demonstrate that models trained on LongCrafter-generated data consistently outperform prior long-context SFT baselines on LongBench~\citep{longbench}, LongBench v2~\citep{longbenchv2}, and LooGLE~\citep{loogle}. On the Qwen2.5-7B~\citep{qwen2.5} and LLaMA-3.1-8B~\citep{llama3} backbones, they achieve the highest average scores across the three benchmarks, reaching \textbf{45.15\%} and \textbf{45.71\%}, respectively, and surpassing the corresponding official post-trained models by \textbf{2.41} and \textbf{5.25} points.
 Further analysis reveals that LongCrafter-trained models robustly identify evidence across different context positions and concentrate attention on the true evidence documents, effectively mitigating the ``lost in the middle'' phenomenon~\citep{lost}.

The main contributions of this paper are as follows:
\begin{itemize}
    \item To enable the controllable synthesis of diverse and challenging long-context tasks, this paper proposes the \textbf{LongCrafter} framework. It defines a broad and comprehensive task space through a hierarchical taxonomy of 32 fine-grained tasks. Its evidence-graph-based synthesis pipeline further organizes task-relevant evidence spans and their cross-paragraph dependencies to guide instruction generation toward the target difficulty and reasoning requirements of each task.

    \item To provide explicit faithfulness supervision, LongCrafter derives responses from the constructed evidence graphs and generates step-by-step reasoning grounded in localized evidence, with verbatim citations to the supporting source spans. Extensive experiments show that LongCrafter produces higher-quality data and enables stronger model performance than existing methods, while effectively mitigating the ``lost in the middle'' problem.
\end{itemize}
%-----------------------------------------------------------------------

%-----------------------------------------------------------------------

\begin{figure}[!t]
    \centering
    \includegraphics[width=0.9\columnwidth]{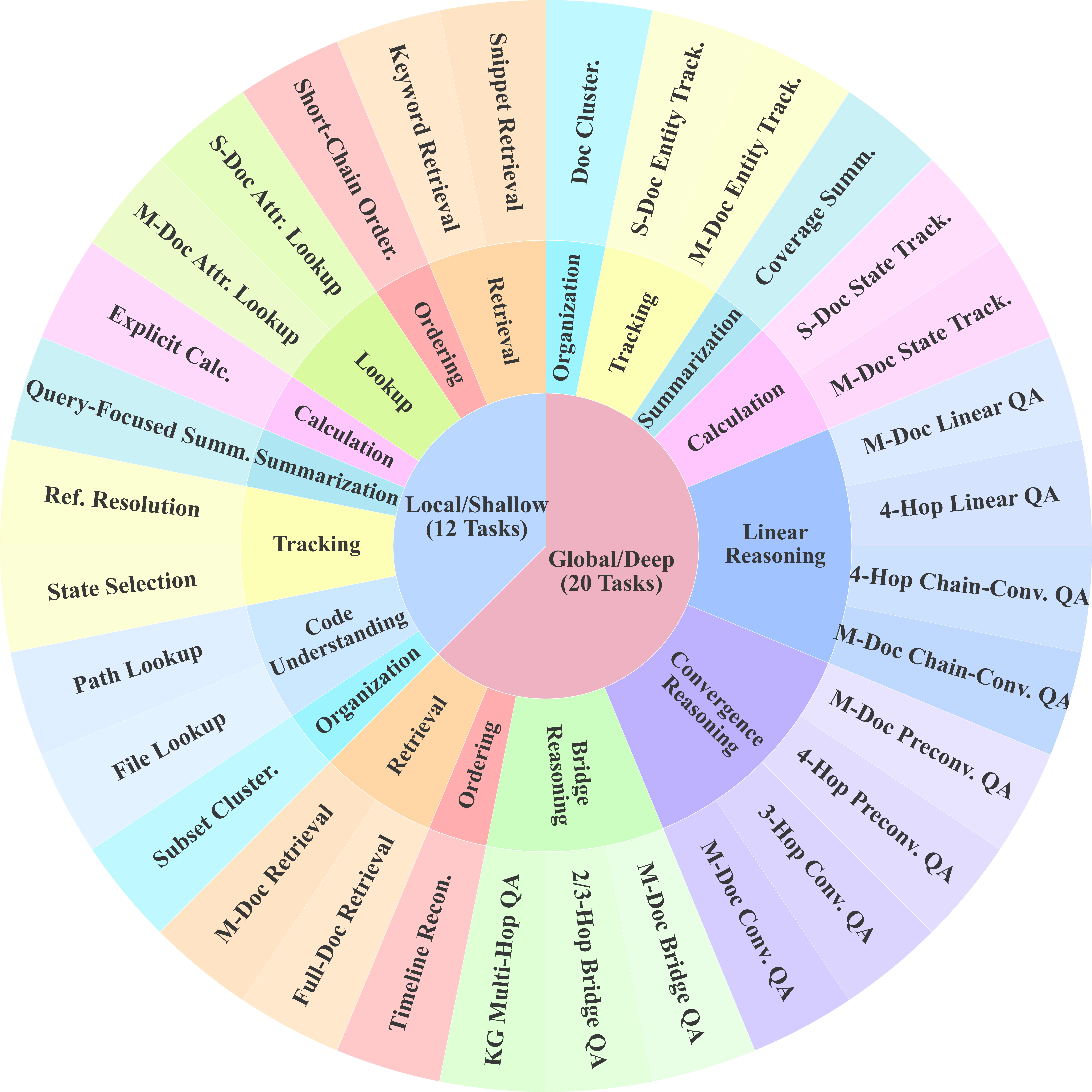}
    \caption{
Hierarchical task taxonomy of LongCrafter.
The inner ring divides long-context understanding into Local/Shallow (12 tasks) and Global/Deep (20 tasks), the middle ring groups them by required capability, and the outer ring lists 32 fine-grained task types.
}
    \label{fig:task}
   
\end{figure}

\begin{figure*}[t]
\centering
\includegraphics[width=1\textwidth]{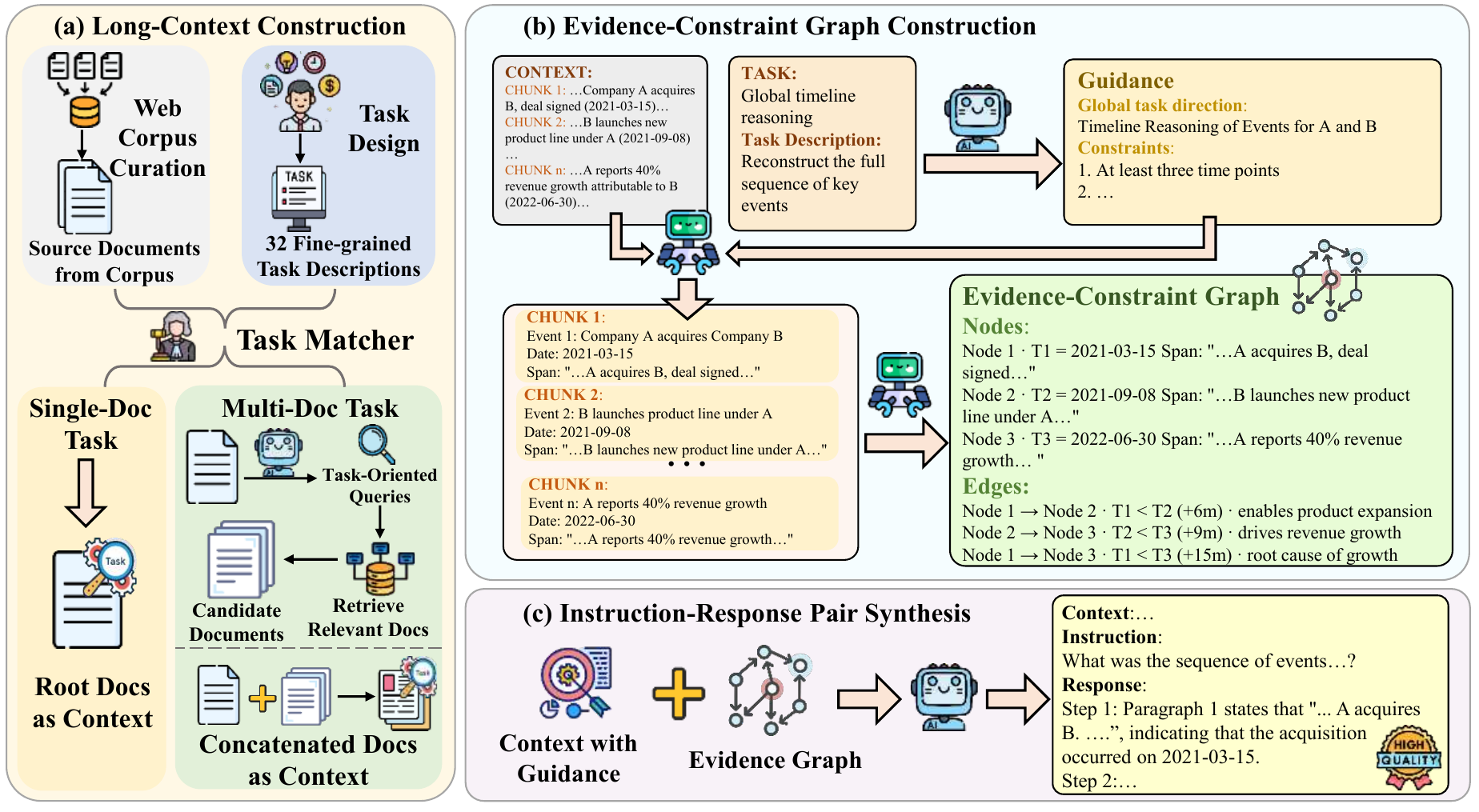}
\caption{Overview of the LongCrafter pipeline. (a) Long-Context Construction: using a curated corpus and 32 fine-grained task types, it matches documents to tasks and builds single- or multi-document contexts; (b) Evidence-Constraint Graph Construction: extracts task-relevant evidence nodes and dependency edges under task-specific guidance; (c) Instruction-Response Pair Synthesis: generates evidence-grounded instructions and citation-based responses from the context and evidence graph.}
\label{fig:pipeline}

\end{figure*}

\section{Methodology}
This section presents the LongCrafter pipeline shown in Figure~\ref{fig:pipeline} for constructing high-quality long-context training data and explains how it addresses the aforementioned limitations.

%% ---------------------------------------------------------------
\subsection{Task Taxonomy}
\label{sec:taxonomy}
%% ---------------------------------------------------------------
LongCrafter builds on a manually constructed task taxonomy grounded in long-context understanding requirements. As shown in Figure~\ref{fig:task}, it first distinguishes tasks that rely on localized evidence from those requiring the integration of globally distributed evidence, and then groups them by required capabilities, yielding 32 fine-grained task types across single- and multi-document settings. This taxonomy broadly covers the long-context understanding capabilities required by existing work and guides all downstream stages.

%% ---------------------------------------------------------------
\subsection{Data Synthesis}
\label{sec:pipeline}
%% ---------------------------------------------------------------
Guided by the task taxonomy, the LongCrafter pipeline produces training samples through three sequential stages: Long-Context Construction, Evidence-Constraint Graph Construction, and Instruction-Response Pair Synthesis. 
%% ---------------------------------------------------------------
\subsubsection{Stage 1: Long Context Construction}
\label{sec:stage1}
To ensure broad domain coverage, we collect and preprocess documents from diverse web sources, forming a rich corpus. We first perform a coarse-grained matching: the corpus is grouped by source, and each source is paired with its candidate fine-grained task types (e.g., code documents are matched to code-understanding tasks), narrowing the search space for the subsequent document-level matching. We then perform document-centric matching, associating each document with applicable task categories along two distinct paths determined by document scope.
\paragraph{Single-document matching.}
Each document is matched to one suitable task from the available single-document categories, such as global summarization or intra-document multi-hop reasoning. The document is then evenly divided into chunks, which serve as the context for Stage~2.

\paragraph{Multi-document matching.}
Each document is also evaluated as a root document for multi-document tasks. If suitable, an LLM generates retrieval queries conditioned on the root document and the target task type, and a hybrid BM25--dense retrieval strategy retrieves complementary documents from the remaining corpus. An LLM then performs joint relevance and diversity filtering to assemble the final context with the root document, keeping the set topically related yet non-redundant---a prerequisite for meaningful cross-document evidence graphs in Stage~2. The assembled context is then split into equal-sized chunks and passed to Stage~2.
%% ---------------------------------------------------------------

\subsubsection{Stage 2: Evidence-Constraint Graph Construction}
\label{sec:stage2}
Stage~2 is the core component that distinguishes LongCrafter from existing methods. Rather than generating questions directly from raw context, LongCrafter first decomposes the context into a structured evidence graph that makes cross-paragraph dependencies explicit. The graph is built via an Extract-then-Construct procedure: exhaustive candidate span extraction followed by task-driven graph construction.

\paragraph{Step 1: Exhaustive Span Extraction.}
Given context $\mathcal{C} = \{c_1, c_2, \ldots, c_n\}$ (where $c_i$ denotes the $i$-th chunk) and target task type $t$, an LLM extracts task-relevant candidate evidence spans from each chunk $c_i$ with reference to the full context. For each candidate, the model records its verbatim snippet, source location, and task-necessary key information. This yields a dense candidate node set $V^{+} = \bigcup_{i=1}^{n} V^{+}_i$, where $V^{+}_i$ denotes all candidates from chunk $c_i$, avoiding premature discarding of potentially necessary evidence that proves crucial only in cross-chunk reasoning.
\paragraph{Step 2: Task-Driven Graph Construction.}
Given the full candidate set $V^{+}$, the model performs a global reasoning pass over the complete context to construct the minimal sufficient evidence graph $G = (V, E)$ with $V \subseteq V^{+}$, where:
\begin{itemize}
    \item Node $v_i \in V$: an evidence span with a paragraph-level citation and task-necessary notes;
    \item Edge $e_{ij} \in E$: a directed cross-paragraph dependency, such as temporal, causal, or coreference relations.
\end{itemize}
Concretely, the model selects the candidate spans jointly necessary and sufficient for task type $t$, and establishes directed dependency edges $E$ between the selected nodes, annotating each with its cross-chunk relation type to capture their logical relationships. A valid evidence graph must satisfy two conditions: (1) the nodes and edges are indispensable for the target task, ensuring the graph faithfully reflects task requirements such as reasoning complexity; (2) the node set uniquely supports the correct answer. Thus, $V$ constrains \textit{which} evidence is needed, while $E$ constrains \textit{how} it is connected, guiding Stage~3 toward inter-span reasoning rather than isolated facts. As a safeguard for global/deep task types, if the selected nodes all originate from the same chunk, targeted re-extraction is triggered to ensure $G$ captures genuine multi-span evidence.

%% ---------------------------------------------------------------
\subsubsection{Stage 3: Instruction-Response Pair Synthesis}
\label{sec:stage3}
Stage~3 generates the final instruction-response pairs conditioned on the evidence graph $G$ produced by Stage~2.
\paragraph{Instruction generation.}
Taking task type $t$, evidence graph $G$, and context $\mathcal{C}$ as input, an LLM generates questions centered on the nodes and edges of $G$, under task-specific constraints (e.g., for global/deep tasks, a correct answer must jointly utilize all key nodes in $G$ and cannot be resolved from any single local passage), thereby ensuring controllable difficulty by construction.
\paragraph{Response generation.}
Responses adopt a step-by-step citation format: each reasoning step locates the corresponding evidence in the source and quotes it verbatim, and the final answer is logically derived from the resulting citation chain, with no external parametric knowledge introduced. This format inherits the node location information and edge dependency ordering from $G$, ensuring end-to-end correspondence between responses and the original context.

The generated pairs are then validated by an LLM that checks whether each instruction is unambiguous, answerable from the given context, and admits a unique correct answer; pairs failing any criterion are discarded. Using GLM-5 throughout the entire pipeline, we generate 2{,}000 high-quality long-context training samples.

\begin{table}[h]
\centering
\begingroup
\small
\makeatletter
\typeout{TABLE FONT SIZE = \f@size pt}
\makeatother
\setlength{\tabcolsep}{1mm}
{
\begin{tabular}{@{}lcccccc@{}}
\toprule
\textbf{Dataset}
  & \textbf{U3G}
  & \textbf{L2}
  & \textbf{Cos.}
  & \textbf{KNN}
  & \textbf{Inertia}
  & \textbf{Radius} \\
\midrule
LongAlign
  & 25{,}670 & 9.70 & 0.24 & 4.55 & 20.31 & 0.23 \\
LongReward
  & 15{,}162 & 7.21 & 0.16 & 3.44 & 13.62 & 0.19 \\
LongMagpie
  & 27{,}737 & 7.98 & 0.15 & 5.36 & 24.30 & 0.20 \\
LongFaith
  & 20{,}116 & 7.51 & 0.14 & 4.60 & 22.84 & 0.19 \\

\textbf{LongCrafter}
  & \textbf{100{,}598}
  & \textbf{10.74}
  & \textbf{0.29}
  & \textbf{5.60}
  & \textbf{32.66}
  & \textbf{0.27} \\
\bottomrule
\end{tabular}
}
\endgroup
\caption{Data diversity across long-context SFT datasets. U3G denotes Unique 3-grams; L2 and Cos. denote mean L2 and cosine distances, respectively. Higher values indicate greater overall diversity.}
\label{tab:data_diversity}
\end{table}

\begin{table*}[ht]
\centering
\begingroup
\small
\makeatletter
\typeout{TABLE FONT SIZE = \f@size pt}
\makeatother
\setlength{\tabcolsep}{1mm}
\begin{tabular}{l|ccccc|ccc|ccccc|c}
\hline
\multicolumn{1}{c|}{\multirow{2}{*}{\textbf{Model}}}
  & \multicolumn{5}{c|}{\textbf{LongBench}}
  & \multicolumn{3}{c|}{\textbf{LongBench v2}}
  & \multicolumn{5}{c|}{\textbf{LooGLE}}
  & \multirow{2}{*}{\textbf{Overall}} \\
\multicolumn{1}{c|}{}
  & \textbf{Qasp} & \textbf{Musq} & \textbf{Wi.QA} & \textbf{Ho.QA} & \textbf{Avg.}
  & \textbf{Easy} & \textbf{Hard} & \textbf{Avg.}
  & \textbf{CR} & \textbf{Comp} & \textbf{TR} & \textbf{MIR} & \textbf{Avg.}
  & \\
\hline

  Qwen2.5-7B-Instruct
  &   {57.5} &   {35.5} &   {60.0} &   {74.0} &   {56.8}
  &   {30.22} &   {27.65} &   {28.63}
  &   {53.0} &   {29.0} &   {53.0} &   {36.0} &   {42.8} &   {42.74} \\
\hdashline

Qwen2.5-7B
  & & & & & & & & & & & & & & \\
\hspace{1.0em}+ LongAlign
  & \underline{51.0} & 29.5 & 51.5 & 63.5 & 48.9
  & \textbf{32.47} & 22.72 & 26.44
  & \underline{51.0} & 24.0 & 38.0 & 27.0 & 35.0 & 36.78 \\
\hspace{1.0em}+ LongReward
  & 50.5 & 38.0 & 54.0 & 67.5 & 52.5
  & 31.25 & \underline{24.33} & \underline{26.97}
  & 45.0 & 24.0 & 38.0 & 29.0 & 34.0 & 37.82 \\
\hspace{1.0em}+ LongMagpie
  & 50.5 & 39.0 & 55.0 & 67.0 & 52.9
  & \underline{31.60} & 23.26 & 26.44
  & 47.0 & \underline{25.0} & \underline{49.0} & \underline{32.0} & \underline{38.3} & 39.21 \\
\hspace{1.0em}+ LongFaith
  & 47.0 & \underline{46.0} & \underline{71.0} & \underline{73.5} & \underline{59.4}
  & 25.69 & \underline{24.33} & 24.85
  & 45.0 & 20.0 & 43.0 & 27.0 & 33.8 & \underline{39.35} \\
\hspace{1.0em}\textbf{+ LongCrafter (Ours)}
  & \textbf{52.5} & \textbf{49.0} & \textbf{74.0} & \textbf{75.5} & \textbf{62.8}
  & 30.03 & \textbf{28.62} & \textbf{29.16}
  & \textbf{53.0} & \textbf{31.0} & \textbf{53.0} & \textbf{37.0} & \textbf{43.5} & \textbf{45.15} \\
\hline

  LLaMA-3.1-8B-Instruct
  &   {53.0} &   {33.0} &   {58.0} &   {69.0} &   {53.25}
  &   {30.73} &   {25.72} &   {27.63}
  &   {51.0} &   {30.0} &   {46.0} &   {35.0} &   {40.5} &   {40.46} \\
\hdashline

LLaMA-3.1-8B
  & & & & & & & & & & & & & & \\
\hspace{1.0em}+ LongAlign
  & 51.0 & 30.5 & 45.0 & 60.0 & 46.6
  & 26.04 & 23.79 & 24.65
  & 52.0 & \underline{29.0} & 37.0 & 39.0 & \underline{39.25} & 36.83 \\
\hspace{1.0em}+ LongReward
  & 50.0 & 38.5 & 45.5 & 63.5 & 49.4
  & 23.96 & 21.97 & 22.73
  & 52.0 & 27.0 & 39.0 & 37.0 & 38.75 & 36.96 \\
\hspace{1.0em}+ LongMagpie
  & \underline{51.5} & 46.0 & 62.5 & 67.5 & 56.9
  & 26.30 & \underline{25.08} & \underline{25.55}
  & \underline{54.0} & 26.0 & 30.0 & \underline{41.0} & 37.75 & 40.07 \\
\hspace{1.0em}+ LongFaith
  & 45.5 & \underline{56.5} & \textbf{71.0} & \underline{69.0} & \underline{60.5}
  & \underline{29.51} & 21.76 & 24.72
  & 52.0 & 21.0 & \underline{40.0} & 31.0 & 36.0 & \underline{40.41} \\
\hspace{1.0em}\textbf{+ LongCrafter (Ours)}
  & \textbf{53.0} & \textbf{57.5} & \underline{70.5} & \textbf{72.5} & \textbf{63.4}
  & \textbf{30.47} & \textbf{25.24} & \textbf{27.24}
  & \textbf{57.0} & \textbf{34.0} & \textbf{53.0} & \textbf{42.0} & \textbf{46.5} & \textbf{45.71} \\
\hline
\end{tabular}
\endgroup
\makeatletter
\typeout{TABLE FONT SIZE = \f@size pt}
\makeatother
\caption{Overall performance (\%) on three benchmarks. CR, Comp, TR, and MIR denote the four LooGLE subsets. Bold and underlined scores mark the best and second-best SFT variants for each backbone.}
\label{tab:main}

\end{table*}

\section{Dataset Analysis}
To assess the quality of LongCrafter-generated training data, we analyze it from multiple perspectives, focusing on its ability to generate diverse high-difficulty instructions at controlled difficulty levels and produce faithful responses.

\subsection{Diversity}
We assess the diversity enabled by our task taxonomy using six metrics: Unique 3-grams~\citep{3grams}, Mean L2 Distance, Mean Cosine Distance, KNN Distance~\citep{distance}, Cluster Inertia~\citep{cluster}, and Radius~\citep{radius}. As shown in Table~\ref{tab:data_diversity}, LongCrafter achieves the best performance on all six, reflecting the broad semantic coverage of our taxonomy-guided construction.
\begin{figure}[!h]
    \centering
    \includegraphics[width=\columnwidth]{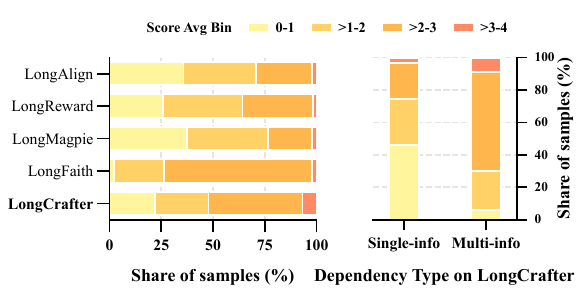}
    \caption{
Instruction difficulty across long-context SFT datasets and different dependency types. Left: distributions over four difficulty bins. Right: LongCrafter samples with single- versus multi-information dependencies. Higher scores indicate greater difficulty.
}
    \label{fig:difficulty_distribution}
\end{figure}
\subsection{Instruction Difficulty Distribution}
To assess LongCrafter's ability to synthesize diverse high-difficulty tasks, we use GPT-5 to score instruction difficulty along three dimensions: evidence locality, required computation or transformation, and distractor strength~\citep{difficulty}, and average them into a composite score. Figure~\ref{fig:difficulty_distribution} (left) shows that LongCrafter produces a more balanced distribution across four difficulty bins and a substantially larger proportion of high-difficulty samples than the baselines. LongFaith, which relies on human-annotated instructions, is concentrated at medium difficulty. Figure~\ref{fig:difficulty_distribution} (right) shows that information-intensive task types receive higher scores, demonstrating our framework’s controllable synthesis at the intended difficulty.

\begin{figure}[!h]
    \centering
    \includegraphics[width=\columnwidth]{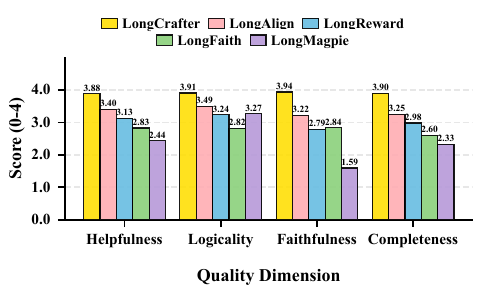}
    \caption{Response quality across training datasets in terms of helpfulness, logicality, faithfulness, and completeness, scored from 0 to 4 (higher is better).}
    \label{fig:response_quality}
\end{figure}

\subsection{Response Quality}
To evaluate LongCrafter’s response quality, particularly faithfulness to the context, we follow the four dimensions defined by~\citet{longreward} and employ an LLM-as-judge protocol to assess Helpfulness, Logicality, Faithfulness, and Completeness. As shown in Figure~\ref{fig:response_quality}, LongCrafter achieves the highest scores across all four dimensions (3.88, 3.91, 3.94, and 3.90), outperforming all baselines. The improvement is especially pronounced in Faithfulness: while the lowest-performing baseline scores only 1.59, LongCrafter achieves 3.94. This result indicates that explicit evidence grounding effectively prevents models from injecting unsupported parametric knowledge into their reasoning chains, thereby promoting faithful, context-grounded reasoning.

\subsection{Human Evaluation}
Following~\citet{human-eval}, we conduct a human--LLM agreement study to verify the reliability of our LLM-as-judge evaluation. We sample 500 instances (100 per dataset), each independently scored by five expert annotators using the same evaluation criteria. We assess inter-annotator reliability using the intraclass correlation coefficient (ICC), obtaining 0.82 for instruction difficulty and 0.87 for response quality, indicating consistent judgments among experts. We then average the five expert scores for each instance to obtain the final human ratings and compare them with the corresponding LLM scores. Spearman's correlations are significant ($p<0.001$) for both dimensions, moderate for instruction difficulty ($\rho=0.63$) and strong for response quality ($\rho=0.74$), indicating strong alignment. Moreover, experts judge that 98\% of LongCrafter instructions satisfy the reasoning-complexity requirement of their assigned task type, confirming that our taxonomy-guided synthesis reliably controls difficulty through task type selection.

%-----------------------------------------------------------------------

\section{Experiments}
%-----------------------------------------------------------------------
In this section, we describe the experimental setup and evaluate the effectiveness of LongCrafter through the performance of models trained on its synthesized data.
\subsection{Experimental Setup}

\paragraph{Model Training.}
We conduct experiments on two representative open-source base models: Qwen2.5-7B~\citep{qwen2.5} and LLaMA-3.1-8B~\citep{llama3}, both supporting a native context length of 128K tokens. For a fair comparison, all models are trained using LoRA with a learning rate of $5\times10^{-5}$ for 2 epochs, with the LoRA rank, alpha, and dropout set to 32, 64, and 0.1, respectively.

\paragraph{Baselines.}
We compare LongCrafter against four representative long-context SFT data construction methods.
LongAlign~\citep{longalign} constructs long instruction-following data from diverse document sources using Self-Instruct.
LongMagpie~\citep{longmagpie} is a self-synthesis framework that leverages aligned LLMs to autoregressively generate contextually relevant instructions given a document.
LongFaith~\citep{longfaith} synthesizes attribution-based reasoning data grounded in the human-annotated MuSiQue dataset.
LongReward~\citep{longreward} filters SFT data by scoring synthesized responses along four dimensions: helpfulness, logicality, faithfulness, and completeness.
Following prior work, we sample 2,000 examples from each dataset for model training to ensure a fair comparison.
In addition, we include the officially released post-trained models of LLaMA and Qwen as strong reference baselines.

\paragraph{Evaluation.}
We evaluate on three long-context benchmarks. \textbf{LongBench}~\citep{longbench} covers single-document and multi-hop reasoning with Qasper, HotpotQA, MuSiQue, and 2WikiMultihopQA. \textbf{LongBench v2}~\citep{longbenchv2} contains 503 challenging multiple-choice questions from realistic scenarios across six task categories. \textbf{LooGLE}~\citep{loogle} evaluates long-dependency QA through Comprehension \& Reasoning, Computation, Timeline Reorder, and Multiple Information Retrieval. We randomly sample 100 examples from each LooGLE subtask. Following prior work, we use GPT-5 to judge answer correctness.
 LongBench v2 results are averaged over three runs with a temperature of 0.1, while all other experiments use a single run with a temperature of 0.

\subsection{Main Results}
\label{sec:main_results}

\paragraph{Overall Performance.}
As shown in Table~\ref{tab:main}, models trained on LongCrafter data attain the highest overall average scores on both backbones (\textbf{45.15\%} on Qwen2.5-7B and \textbf{45.71\%} on LLaMA-3.1-8B), ranking first on the large majority of individual subtasks and surpassing the second-best SFT baseline by \textbf{+5.80} and \textbf{+5.30} points, respectively. This cross-architecture consistency indicates that the gains stem from the data construction design itself rather than any backbone-specific inductive bias, and that LongCrafter's broad task diversity yields strong generalization rather than overfitting to a particular task distribution.

\paragraph{Gains on Difficult Tasks.}
On LongBench~v2, which targets deeper reasoning over realistic long contexts, LongCrafter achieves the best overall results among all SFT variants, reaching \textbf{29.16\%} on Qwen2.5-7B and \textbf{27.24\%} on LLaMA-3.1-8B. On LooGLE, whose tasks require integrating evidence dispersed across multiple locations, it leads across all four subtasks, with an especially large margin on Timeline Reorder on LLaMA, where it scores \textbf{53.0\%} against 40.0\% for the second-best SFT baseline.

\paragraph{Efficiency.}
With only 2{,}000 training samples, LongCrafter outperforms the officially post-trained models on the Overall metric, with gains of \textbf{+2.41} and \textbf{+5.25} points on Qwen2.5-7B and LLaMA-3.1-8B, respectively. This suggests that carefully designed, evidence-grounded data can drive substantial long-context gains at minimal data scale.

\subsection{Position-Robust Evidence Localization}

\begin{figure}[h]
    \centering
    \includegraphics[width=\linewidth]{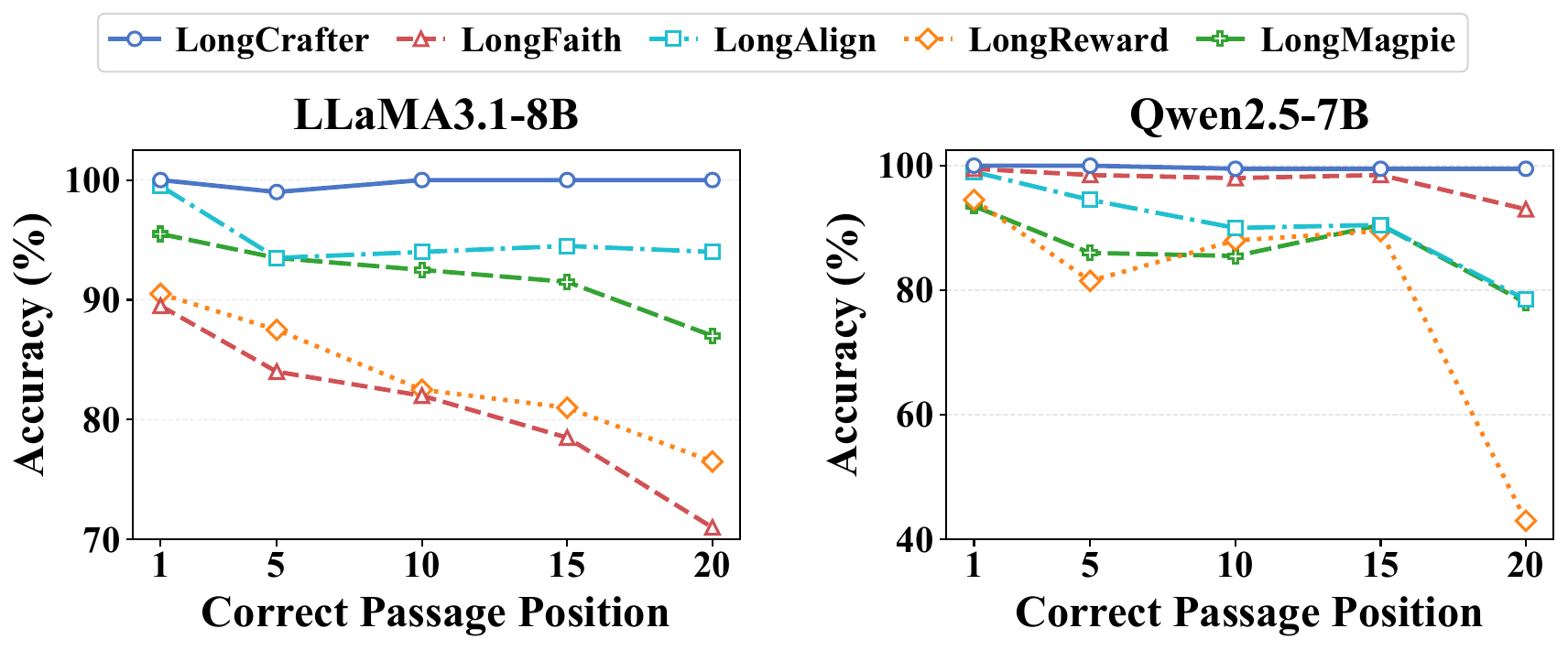}
 
    \caption{
Evidence-position robustness across training datasets. Retrieval accuracy is reported with the gold document placed at positions 1, 5, 10, 15, or 20.
}
    \label{fig:attention_score}
   
\end{figure}

\paragraph{Evidence Position Robustness.}
To test whether models trained on our data learn content-based evidence localization rather than positional shortcuts, we follow~\citet{never-lost} and place the gold evidence document at positions 1, 5, 10, 15, and 20 among 20 candidates, probing the ``lost in the middle'' problem~\citep{lost}. As shown in Figure~\ref{fig:attention_score}, LongCrafter-trained models maintain near-perfect retrieval accuracy across all positions on both backbones, whereas the baselines decline as the evidence moves later in the context, dropping to about 71\% for LongFaith on LLaMA-3.1-8B and 40\% for LongReward on Qwen2.5-7B.

\begin{figure}[h]
    \centering
    \includegraphics[width=\linewidth]{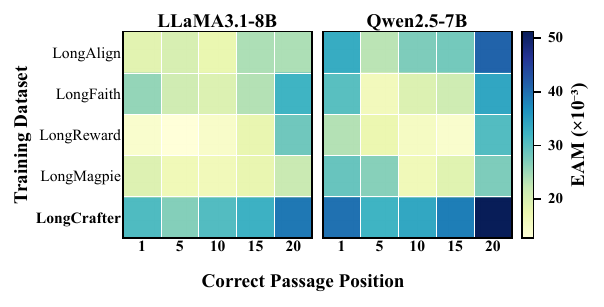}
 
    \caption{
Evidence Attention Mass (EAM) on gold evidence tokens across training datasets and passage positions. Rows indicate datasets, columns indicate evidence positions, and darker cells denote higher EAM.}
    \label{fig:eam}

\end{figure}

\paragraph{Attention Mass Analysis.}
Following~\citet{EAM}, we measure how much attention the model assigns to gold evidence tokens. For a response with $T$ generated tokens, the Evidence Attention Mass (EAM) for layer $\ell$ and head $h$ is:
\[
  \overline{\mathrm{EAM}}^{(\ell,h)} = \frac{1}{T} \sum_{t=1}^{T} \sum_{j \in \mathcal{E}_t} a^{(\ell,h)}_{t,j},
\]
where $a^{(\ell,h)}_{t,j}$ is the attention probability to evidence token $j$ at decoding step $t$, and $\mathcal{E}_t$ contains visible gold evidence tokens. Averaged over layers, heads, and samples, Figure~\ref{fig:eam} shows that LongCrafter-trained models achieve the highest attention mass across positions on both backbones, even at position~20 where baselines are weakest. 

Together, these results show that our evidence-graph-guided training consistently improves evidence position robustness, enabling models to locate and attend to relevant evidence regardless of its position.

\subsection{Ablation}

We conduct ablation studies from two perspectives: component ablations and data distribution ablations. The results are shown in Table~\ref{tab:ablation}.

\label{sec:ablation}

\paragraph{Component Ablations.}
We first ablate LongCrafter's core components, including Evidence Graph Construction (EGC) and Evidence-Based Citation (EBC). Removing EGC substantially lowers the difficulty and quality of synthesized data: the average difficulty score of Global/Deep tasks drops from 2.59 to 1.74. To further assess data quality, we manually inspect 100 samples generated without EGC and find that about 30\% are clearly low-quality, including insufficiently challenging instructions, answers that are not faithful to the original context, or flawed reasoning logic due to the absence of explicit evidence-graph edges that support logical dependencies among evidence spans.

The downstream results show the same trend. Removing EGC causes the largest performance drop, reducing the average score by 12.64 points on Qwen2.5-7B and 11.89 points on LLaMA-3.1-8B. Removing EBC also consistently hurts performance, with average drops of 5.25 points and 4.08 points, respectively. These results indicate that EGC improves both the difficulty and quality of generated instructions, while providing reliable evidence dependencies for synthesizing high-quality answers. EBC further strengthens faithful supervision, encouraging models to reason based on the original evidence.

\paragraph{Data Distribution Ablations.}
We further study the effect of data distribution by comparing \textit{Easy-only}, \textit{Hard-only}, and \textit{Low-diversity} settings. The \textit{Easy-only} setting samples only local-information-dependent tasks, while the \textit{Hard-only} setting samples only global-information-dependent tasks. Both underperform the full LongCrafter setting, with average drops of 8.29/6.80 points and 7.24/5.15 points on Qwen2.5-7B/LLaMA-3.1-8B, respectively. The \textit{Low-diversity} setting, which samples only from common base task types, also reduces the average score by 7.72/6.78 points. These results indicate that robust long-context generalization is best achieved by diverse task types and balanced difficulty.

Overall, the proposed task taxonomy ensures balanced and diverse task distributions for better long-context generalization, while EGC and EBC ensure instruction difficulty, data quality, and explicit faithfulness supervision.

\begin{table}[t]
\centering
\begingroup
\small
\makeatletter
\typeout{TABLE FONT SIZE = \f@size pt}
\makeatother
\setlength{\tabcolsep}{1mm}
\begin{tabular}{@{}lccccc@{}}
\toprule
\textbf{Model}
  & \textbf{LBv1}
  & \textbf{LBv2}
  & \textbf{LooGLE}
  & \textbf{Avg.}
  & \textbf{$\Delta$} \\
\midrule

\multicolumn{6}{@{}l}{Qwen2.5-7B} \\
\textbf{LongCrafter}
  & \textbf{62.8} & \textbf{29.16} & \textbf{43.5} & \textbf{45.15} & -- \\

\addlinespace[1pt]
\multicolumn{6}{@{}l}{\textbf{Component}} \\
\quad w/o EGC
  & 41.5 & 22.53 & 33.5 & 32.51
  & {$-12.64$} \\
\quad w/o EBC
  & 60.3 & 22.40 & 37.0 & 39.90
  & {$-5.25$} \\

\addlinespace[1pt]
\multicolumn{6}{@{}l}{\textbf{Data Dist.}} \\
\quad Easy-only
  & 57.9 & 18.89 & 33.8 & 36.86
  & {$-8.29$} \\
\quad Hard-only
  & 60.6 & 18.62 & 34.5 & 37.91
  & {$-7.24$} \\
\quad Low-div.
  & 57.5 & 19.48 & 35.3 & 37.43
  & {$-7.72$} \\

\midrule

\multicolumn{6}{@{}l}{LLaMA-3.1-8B} \\
\textbf{LongCrafter}
  & \textbf{63.4} & \textbf{27.24} & \textbf{46.5} & \textbf{45.71} & -- \\

\addlinespace[1pt]
\multicolumn{6}{@{}l}{\textbf{Component}} \\
\quad w/o EGC
  & 43.2 & 23.26 & 35.0 & 33.82
  & {$-11.89$} \\
\quad w/o EBC
  & 62.1 & 21.80 & 41.0 & 41.63
  & {$-4.08$} \\

\addlinespace[1pt]
\multicolumn{6}{@{}l}{\textbf{Data Dist.}} \\
\quad Easy-only
  & 57.4 & 19.82 & 39.5 & 38.91
  & {$-6.80$} \\
\quad Hard-only
  & 59.5 & 19.88 & 42.3 & 40.56
  & {$-5.15$} \\
\quad Low-div.
  & 58.6 & 19.88 & 38.3 & 38.93
  & {$-6.78$} \\

\bottomrule
\end{tabular}
\endgroup
\makeatletter
\typeout{TABLE FONT SIZE = \f@size pt}
\makeatother
\caption{Component and data distribution ablations for LongCrafter on long-context benchmarks.}
\label{tab:ablation}
\end{table}
\begin{figure}[h]
    \centering
    \includegraphics[
        width=\linewidth,
    ]{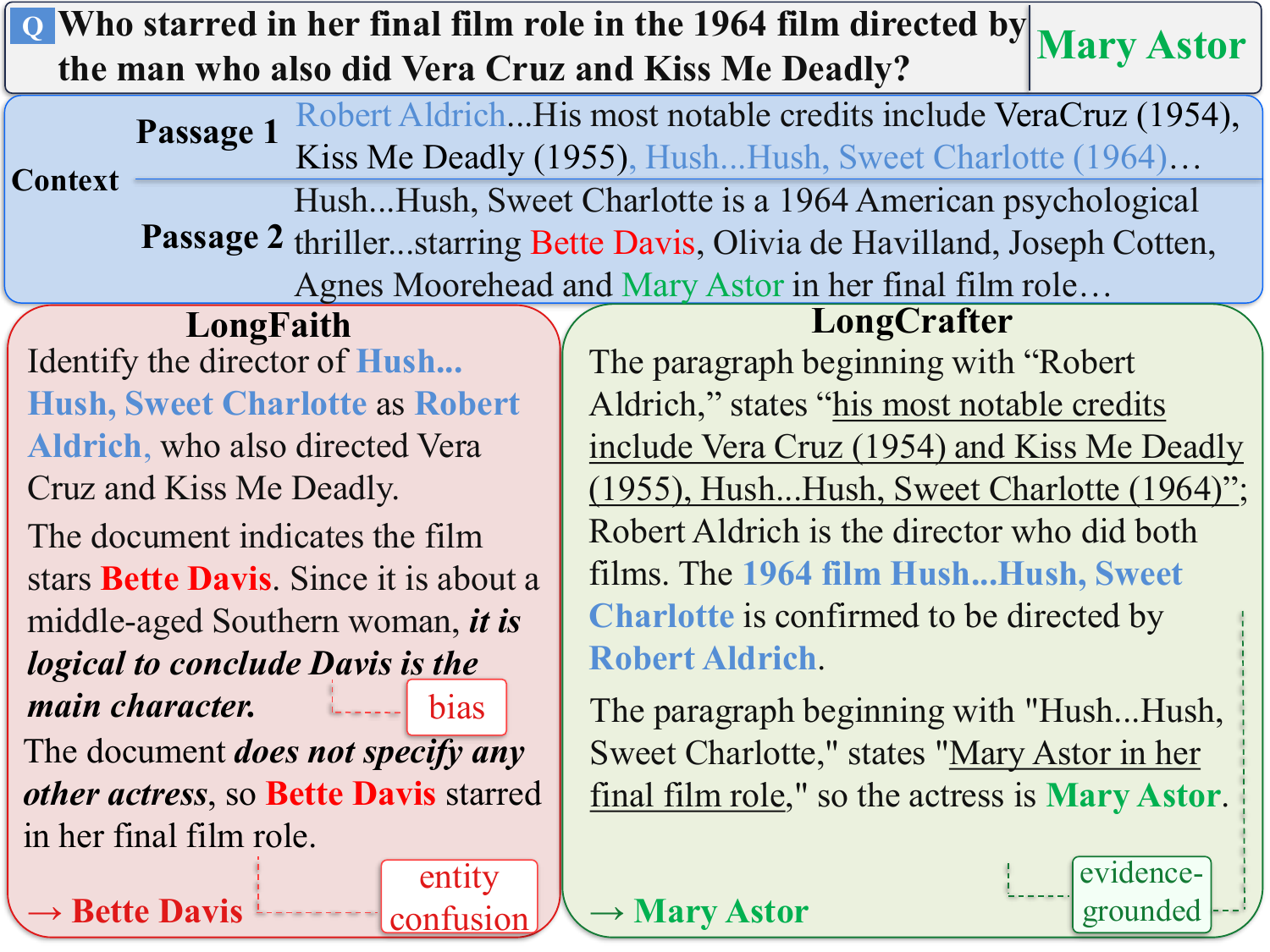}
    \caption{
    Case study comparing LongCrafter with a baseline (LongFaith) on an evidence-grounded multi-hop question.
    }
    \label{fig:case_study}
    
\end{figure}
\subsection{Case Study}

Figure~\ref{fig:case_study} illustrates how evidence-grounded reasoning improves answer faithfulness. The question requires a three-hop chain: identifying the director of \textit{Vera Cruz} and \textit{Kiss Me Deadly}, locating his 1964 film, and naming the actress who made her final film appearance in it. LongFaith identifies the correct film but selects the co-star Bette Davis, confusing the most salient entity with the evidence-supported answer. In contrast, the LongCrafter-trained model grounds each step in explicitly located and cited evidence, correctly tracing Robert Aldrich to \textit{Hush\dots Hush, Sweet Charlotte} and finally to Mary Astor, thereby avoiding unsupported entity bias.

\section{Related Work}
\paragraph{Long-Context Understanding Enhancement.}
Existing work improves long-context understanding through position encoding extensions, including rotary position embedding variants~\citep{yarn,longrope} and position interpolation~\citep{extending}, continued pre-training on long documents~\citep{effective,dataengineering}, and post-training with SFT~\citep{longalign,longlora,longmit}, reinforcement learning~\citep{longreward,rl}, or preference optimization~\citep{dpo,orpo}. Our work focuses on synthesizing long-context SFT data with broad task coverage and faithful evidence grounding.

\paragraph{Long-Context Training Data Synthesis.}
Prior synthesis methods mainly emphasize either \textit{instruction diversity} or \textit{response quality}. Diversity-oriented methods use direct~\citep{longlora,longalign,longmagpie,wildlong}, multi-agent~\citep{longmit,agent}, or graph-based multi-hop synthesis~\citep{cgmis}, but often lack systematic control over task type, difficulty, and faithfulness. Quality-oriented methods improve faithfulness and traceability through coarse-grained evidence attribution~\citep{longfaith,chain} or preference signals~\citep{longreward}, but typically impose evidence constraints only after question generation. In contrast, LongCrafter constructs an explicit evidence graph before instruction generation, enabling diverse task coverage, controlled instruction difficulty, and reasoning grounded in located and cited evidence.
\section{Conclusion}
In this paper, we propose LongCrafter, a structured data synthesis framework for long-context SFT that systematically addresses two key limitations of existing approaches: the lack of methods for controllably synthesizing a diverse range of tasks and the absence of explicit faithfulness supervision. It combines a comprehensive task taxonomy, evidence graphs capturing cross-paragraph dependencies, and responses grounded in explicitly located and cited evidence to produce diverse, difficulty-controlled, and faithful training data. Experiments across three benchmarks and two model families demonstrate consistently strong overall performance, particularly on high-difficulty tasks, while ablations also confirm the effectiveness of the evidence graph and explicit faithfulness supervision.

\bibliography{aaai2026}

\clearpage
\section*{Appendix}
\setcounter{figure}{0}
\setcounter{table}{0}
\renewcommand{\thefigure}{A.\arabic{figure}}
\renewcommand{\thetable}{A.\arabic{table}}
\subsection*{Additional Data Construction Details}
\label{subsec:data_construction}
\paragraph{Corpus Collection.}
We construct a cross-domain bilingual long-context corpus spanning 
11 domains: dialogue, academic papers, structured data, source code, 
judicial documents, legal statutes, news, fiction, scripts, biographies, 
and knowledge graphs. Data are sourced from publicly available 
collections, including WildChat, arXiv, ChinaXiv, GitHub, Project 
Gutenberg, Wikipedia, and Wikidata, among others. Each domain 
naturally exhibits distinct long-context structures, such as 
hierarchical provisions in legal texts, cross-file dependencies in 
code, and entity-relation networks in knowledge graphs, providing diverse training contexts. 
\paragraph{Filtering and Cleaning.}
We apply source-level and sample-level filtering to ensure data 
quality. Source-level filtering removes documents with failed parsing, 
encoding errors, unclear provenance, or duplicate content. Sample-level 
filtering requires at least 50\% target-language characters and 50\% 
non-empty line deduplication rate, and discards texts dominated by 
footnotes, numbered lists, or web noise. All documents are converted 
to Markdown or plain-text format using source-specific parsers, 
followed by domain-specific cleaning rules.

\paragraph{Long-context Sample Construction.}
We construct long-context samples by applying length thresholds: 
single-document samples require at least 5,000 Chinese or 15,000 
English characters, while multi-document samples are assembled within 
controlled length ranges per domain. Overlong documents are truncated 
at paragraph boundaries, and no document is reused across the 
constructed data. Table~\ref{tab} reports the average input and output lengths of the constructed samples and baseline datasets. The hierarchical task taxonomy used to organize fine-grained task types is shown in Table~\ref{tab:task_taxonomy}.

\begin{table}[h]

\centering

\begin{tabular}{lrr}

\toprule

\textbf{Datasets} & \textbf{Input} & \textbf{Output} \\

\midrule

LongMagpie  & 5361.1  & 726.1 \\

LongReward  & 22696.5 & 289.7 \\ 

LongFaith   & 2738.4  & 224.2 \\

LongAlign   & 16465.4 & 186.4 \\ 

LongCrafter & 10269.0 & 386.8 \\

\bottomrule

\end{tabular}
\caption{Average input and output token lengths of baseline datasets and LongCrafter in the main experiments. All datasets contain 2K examples.}
\label{tab}
\end{table}

\begin{figure}[h]
    \centering
    \includegraphics[width=1\columnwidth]{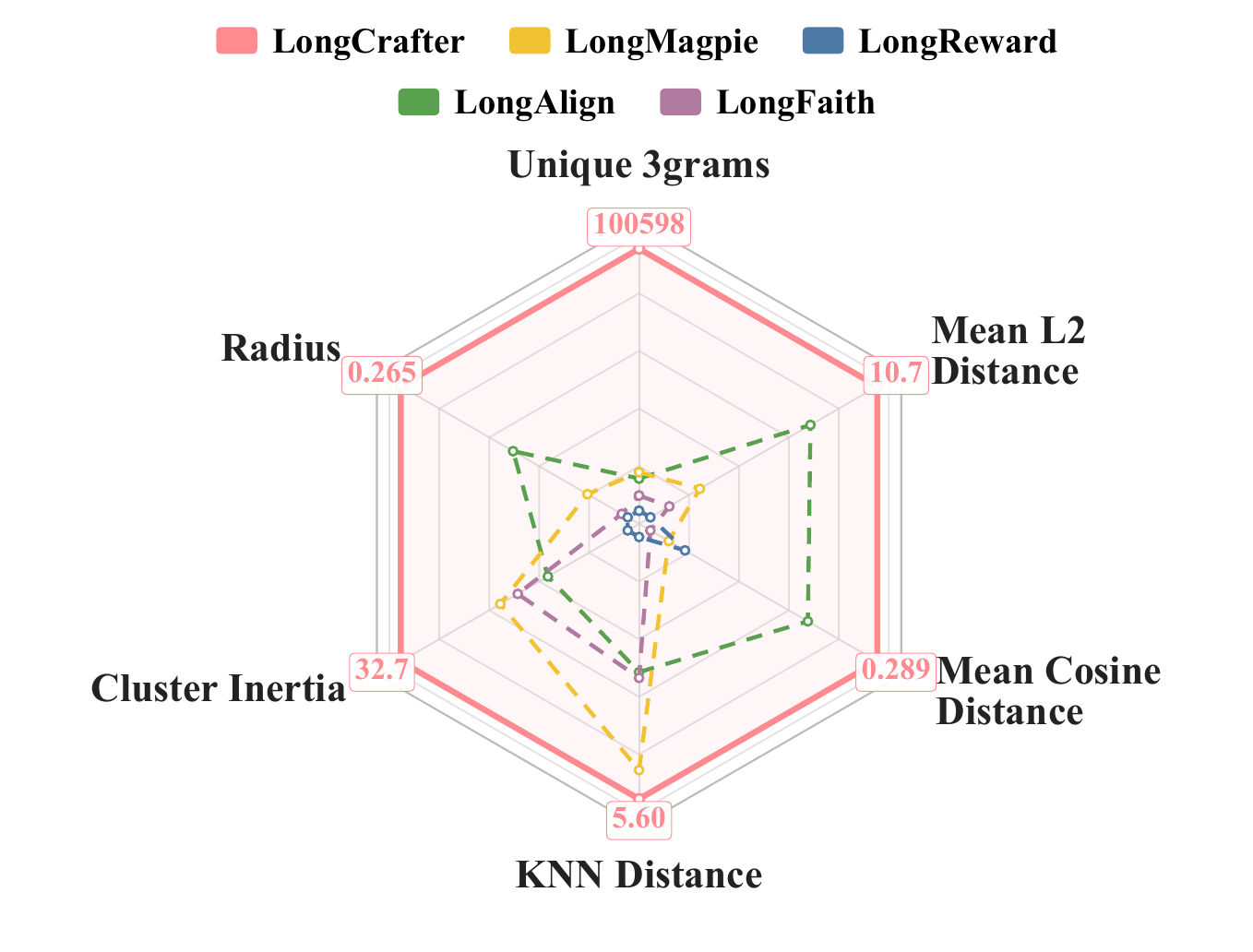}
    \caption{Diversity comparison across different datasets.}
    \label{fig:radar}
   
\end{figure}

\subsection*{Diversity}
\label{app:diversity}

Figure~\ref{fig:radar} reports the diversity comparison among different training datasets from multiple complementary perspectives. These metrics jointly measure lexical variation, global semantic dispersion, local neighborhood separation, and the breadth of the embedding-space distribution. Together, they provide a comprehensive view of how diverse each dataset is in both surface form and semantic coverage.
For a fair comparison, all metrics are computed under the same sampling size and embedding configuration.Higher values indicate broader lexical or semantic coverage. 

\subsection*{Further Evaluation Details}
\label{subsec:evaluation}

 To prevent data contamination, we first applied MinHash-based deduplication to filter out training samples with high similarity to the evaluation sets, followed by manual review to further ensure no data leakage between the training and test sets.
 Following prior work, we use GPT-5 to judge answer correctness. A manual spot-check shows that GPT-5's judgments reach 99.5\% agreement with human verification.

\begin{table*}[t]
\centering
\small
\setlength{\tabcolsep}{3pt}
\renewcommand{\arraystretch}{1.06}

\newcolumntype{L}[1]{>{\raggedright\arraybackslash}p{#1}}
\newcolumntype{Y}{>{\raggedright\arraybackslash}X}

% Soft full-width separator between capability blocks.
\newcommand{\softfullrule}{%
  \addlinespace[1pt]
  \arrayrulecolor{black!30}%
  \cmidrule[\lightrulewidth]{1-4}%
  \arrayrulecolor{black}%
  \addlinespace[1pt]
}

% Soft right-side-only separator for the final Global/Deep block.
\newcommand{\softrightrule}{%
  \addlinespace[1pt]
  \arrayrulecolor{black!30}%
  \cmidrule[\lightrulewidth]{3-4}%
  \arrayrulecolor{black}%
  \addlinespace[1pt]
}

\begin{tabularx}{\textwidth}{@{}L{3.05cm}Y@{\hspace{0.7em}}L{3.65cm}Y@{}}
\toprule
\multicolumn{2}{c}{\textbf{Local/Shallow}} &
\multicolumn{2}{c}{\textbf{Global/Deep}} \\
\cmidrule(r{0.7em}){1-2}
\cmidrule(l{0.7em}){3-4}
\textbf{Capability} & \textbf{Fine-grained Task Types} &
\textbf{Capability} & \textbf{Fine-grained Task Types} \\
\midrule

\textbf{Retrieval}
& snippet retrieval
& \textbf{Retrieval}
& multi-doc retrieval \\
& keyword retrieval
&
& full-doc retrieval \\

\softfullrule

\textbf{Ordering}
& short-chain ordering
& \textbf{Ordering}
& timeline reconstruction \\

\softfullrule

\textbf{Lookup}
& single-doc attribute lookup
& \textbf{Bridge Reasoning}
& KG multi-hop QA \\
& multi-doc attribute lookup
&
& 2/3-hop bridge QA \\
&
&
& multi-doc bridge QA \\

\softfullrule

\textbf{Calculation}
& explicit calculation
& \textbf{Convergence Reasoning}
& multi-doc convergence QA \\
&
&
& 3-hop convergence QA \\
&
&
& 4-hop preconvergence QA \\
&
&
& multi-doc preconvergence QA \\

\softfullrule

\textbf{Summarization}
& query-focused summary
& \textbf{Linear Reasoning}
& multi-doc chain-convergence QA \\
&
&
& 4-hop chain-convergence QA \\
&
&
& 4-hop linear QA \\
&
&
& multi-doc linear QA \\

\softfullrule

\textbf{Tracking}
& reference resolution
& \textbf{Calculation}
& single-doc state tracking \\
& state selection
&
& multi-doc state tracking \\

\softfullrule

\textbf{Code Understanding}
& path lookup
& \textbf{Summarization}
& coverage summary \\
& file lookup
&
& \\

\softfullrule

\textbf{Organization}
& subset clustering
& \textbf{Tracking}
& single-doc entity tracking \\
&
&
& multi-doc entity tracking \\

\softrightrule

&
&
\textbf{Organization}
& doc clustering \\

\bottomrule
\end{tabularx}

\caption{Hierarchical task taxonomy of LongCrafter. Fine-grained task types are grouped by their required capability and organized into local/shallow and global/deep levels of long-context understanding.}
\label{tab:task_taxonomy}
\end{table*}

\subsection*{Prompts}
\label{app:evaluation-prompts}

This section enumerates all prompts utilized in our evaluation framework.
Prompt 1 presents the response quality evaluation
prompt for long-context prompt-response data. It evaluates each response along four dimensions: Helpfulness, Logicality,
Faithfulness, and Completeness.
Prompt 2 describes the instruction difficulty
estimation prompt. It measures difficulty from
three dimensions: evidence\_locality, computation\_or\_transformation, and
distractor\_strength.
Prompt 3 presents the answer correctness judge used for evaluation.

\begin{promptbox}[Prompt 1: Answer Quality Evaluator]
\label{prompt:answer-quality}
You are an impartial and strict evaluator for long-context prompt-response data.

Your task is to evaluate the quality of a model response given the original prompt and the model response. The prompt may contain a long context, documents, tables, instructions, constraints, candidate options, or a question. You must judge the response only based on the information available in the prompt and the response. Do not use external knowledge unless the prompt explicitly asks the model to use general knowledge.

You must evaluate the response from four dimensions:
\begin{enumerate}
    \item Helpfulness
    \item Logicality
    \item Faithfulness
    \item Completeness
\end{enumerate}

Each dimension must receive a score from 0 to 4. You may use integer or one-decimal scores. Higher is better. The final score is the arithmetic mean of the four dimension scores.

General judging principles:
\begin{itemize}
    \item Be strict but fair.
    \item Penalize unsupported claims, fabricated details, missing required constraints, incorrect reasoning, and incomplete answers.
    \item Do not reward verbosity by itself. A long answer is good only if it is relevant, correct, faithful, and complete.
    \item Do not penalize concise answers if they fully satisfy the prompt.
    \item If the response refuses to answer, evaluate whether the refusal is justified by the prompt. Unjustified refusal should receive low helpfulness and completeness.
    \item If the prompt requires a specific output format, candidate selection, ranking, citation, JSON, calculation, or step-by-step result, evaluate whether the response follows that requirement.
    \item If the prompt contains context or source material, treat that context as the primary ground truth.
    \item If the prompt does not contain enough evidence to verify a factual claim, do not assume the claim is true. Mark it as unsupported under faithfulness.
    \item If the prompt contains multiple tasks, constraints, or sub-questions, evaluate all of them.
\end{itemize}

Dimension definitions and scoring criteria:

1. Helpfulness

Evaluate whether the response directly satisfies the user's request.

Consider:
\begin{itemize}
    \item Is the response relevant to the prompt?
    \item Does it answer the actual question or solve the requested task?
    \item Does it follow the user's stated constraints and formatting requirements?
    \item Is it informative enough for the user's purpose?
    \item Does it avoid unnecessary digression?
\end{itemize}

Scoring guide:
\begin{itemize}
    \item 4: Fully addresses the prompt, satisfies all major requirements, and is directly useful.
    \item 3: Mostly helpful, with minor omissions or minor format issues.
    \item 2: Partially helpful, but misses important requirements or answers only part of the task.
    \item 1: Mostly unhelpful, off-target, or fails to follow core instructions.
    \item 0: Completely irrelevant, empty, or refuses without justification.
\end{itemize}

2. Logicality

Evaluate whether the response is internally coherent and logically sound.

Consider:
\begin{itemize}
    \item Are different parts of the response mutually consistent?
    \item Are the reasoning steps, comparisons, calculations, or conclusions valid?
    \item Does the response contradict itself?
    \item Does it make unjustified jumps from premises to conclusions?
    \item If the task involves ordering, classification, aggregation, or multi-hop reasoning, is the reasoning structure valid?
\end{itemize}

Important:
\begin{itemize}
    \item Logicality concerns the internal consistency and reasoning quality of the response.
    \item Do not use external knowledge to judge logicality.
    \item A response can be logically coherent but factually unsupported; that should be penalized under faithfulness, not necessarily under logicality.
\end{itemize}

Scoring guide:
\begin{itemize}
    \item 4: No noticeable logical errors; reasoning is coherent and conclusions follow.
    \item 3: Mostly logical, with only minor unclear or weakly justified steps.
    \item 2: Some valid reasoning, but also noticeable inconsistencies, gaps, or calculation errors.
    \item 1: Major logical contradictions or invalid reasoning.
    \item 0: Incoherent, self-contradictory, or impossible to follow.
\end{itemize}

3. Faithfulness

Evaluate whether the factual content of the response is supported by the prompt, especially by the provided context or documents.

Use the following process mentally:
\begin{itemize}
    \item Identify factual claims in the response.
    \item Ignore purely functional or transitional sentences such as ``In summary'' or ``The answer is as follows'' unless they contain factual claims.
    \item For each factual claim, check whether it is fully supported, partially supported, contradicted, or not supported by the prompt.
    \item Penalize hallucinations, fabricated entities, fabricated numbers, fabricated citations, fabricated causal relations, and claims that go beyond the prompt without permission.
\end{itemize}

Support levels:
\begin{itemize}
    \item Fully supported: The claim is directly stated in or clearly entailed by the prompt.
    \item Partially supported: Some parts are supported, but some details are missing, overstated, or uncertain.
    \item Not supported: The claim is absent from the prompt, contradicted by the prompt, or depends on outside knowledge not allowed by the prompt.
\end{itemize}

Scoring guide:
\begin{itemize}
    \item 4: All or almost all factual claims are supported by the prompt.
    \item 3: Most factual claims are supported; only minor unsupported or slightly overstated details.
    \item 2: Mixed faithfulness; several important claims are unsupported or only partially supported.
    \item 1: Many claims are unsupported, fabricated, or contradicted by the prompt.
    \item 0: The response is largely hallucinated or contradicts the prompt.
\end{itemize}

Special cases:
\begin{itemize}
    \item If the response contains no factual claims but also does not answer the task, faithfulness may be high or neutral, but helpfulness and completeness should be low.
    \item If the prompt lacks source context needed to verify the response, assign a lower faithfulness score when the response makes specific factual claims that cannot be verified from the prompt.
    \item If the task asks for creative writing, judge faithfulness mainly by whether the response respects the user's constraints and does not contradict provided facts.
\end{itemize}

4. Completeness

Evaluate whether the response covers all key information, sub-questions, constraints, and required aspects in the prompt.

Consider:
\begin{itemize}
    \item Does it answer every part of the prompt?
    \item Does it include all required entities, steps, options, comparisons, calculations, evidence, or conclusions?
    \item Does it omit important information from the prompt that is necessary for a satisfactory answer?
    \item In long-context tasks, does it avoid focusing only on early or salient parts while ignoring relevant middle or later information?
    \item If the prompt asks for a ranking, selection, extraction, or structured output, does the response include all required items and no unjustified omissions?
\end{itemize}

Scoring guide:
\begin{itemize}
    \item 4: Fully complete; covers all key points and required aspects.
    \item 3: Mostly complete; only minor omissions.
    \item 2: Partially complete; covers some important points but misses others.
    \item 1: Severely incomplete; misses most required content.
    \item 0: Does not provide the requested answer or omits essentially everything important.
\end{itemize}

Output requirements:

You must output only a valid JSON object. Do not output markdown, explanations outside JSON, or extra text.

The JSON object must follow this schema exactly:

\begin{verbatim}
{
  "helpfulness": {
    "score": number,
    "reason": "brief reason"
  },
  "logicality": {
    "score": number,
    "reason": "brief reason"
  },
  "faithfulness": {
    "score": number,
    "reason": "brief reason"
  },
  "completeness": {
    "score": number,
    "reason": "brief reason"
  }
}
\end{verbatim}

Rules for the JSON:
\begin{itemize}
    \item All scores must be numbers between 0 and 4.
    \item The reasons must be concise but specific.
\end{itemize}
\end{promptbox}

\begin{promptbox}[Prompt 2: Difficulty Evaluator]
\label{prompt:difficulty}
You are a strict and consistent difficulty evaluator for long-context questions.

You will receive one data instance containing:
\begin{itemize}
    \item prompt: the original task prompt given to a model. The prompt is constrained to contain one long context followed by one question. The long context may include passages, documents, tables, records, candidate options, and output-format requirements that are part of the question.
\end{itemize}

Your task is to estimate the intrinsic difficulty of this data instance based only on the prompt.

Difficulty dimensions:
\begin{itemize}
    \item evidence\_locality: Whether the evidence needed to answer the question is local and explicit or dispersed across multiple parts of the long context. Higher means the required evidence appears more scattered and is harder to locate.
    \item computation\_or\_transformation: Whether answering the question requires calculation, aggregation, sorting, comparison, normalization, restructuring, format conversion, or other nontrivial transformation of context information.
    \item distractor\_strength: Whether the long context contains misleading, similar, competing, redundant, or easily confusable evidence/options.
\end{itemize}

Important principles:
\begin{enumerate}
    \item Treat the prompt as a long-context question-answering instance: identify the long context and the single question, then use only them to infer what kind of evidence selection, reasoning, comparison, calculation, aggregation, ordering, or transformation is required.
    \item The main question is: how difficult would it be for a capable LLM to answer the question from the long context?
    \item If the question asks for a short answer but requires dispersed evidence, aggregation, ordering, calculation, or rejecting confusing distractors, the relevant dimensions should still receive high scores.
    \item If the question can be answered by directly copying a local span from the long context, the relevant dimensions should receive low scores.
    \item Do not use external knowledge unless the prompt explicitly requires it.
\end{enumerate}

Dimension score scale:

For each dimension, assign an integer from 0 to 4:
\begin{itemize}
    \item 0: not present / irrelevant
    \item 1: very low
    \item 2: low to moderate
    \item 3: high
    \item 4: very high
\end{itemize}

Output valid JSON only. Do not output markdown, comments, or any text outside the JSON.

Required JSON schema:

\begingroup
\small
\ttfamily
\noindent\{\\
\hspace*{1em}"evidence\_locality": \{\\
\hspace*{2em}"score": <integer from 0 to 4>,\\
\hspace*{2em}"reason": "<concise reason for this dimension>"\\
\hspace*{1em}\},\\
\hspace*{1em}"computation\_or\_transformation": \{\\
\hspace*{2em}"score": <integer from 0 to 4>,\\
\hspace*{2em}"reason": "<concise reason for this dimension>"\\
\hspace*{1em}\},\\
\hspace*{1em}"distractor\_strength": \{\\
\hspace*{2em}"score": <integer from 0 to 4>,\\
\hspace*{2em}"reason": "<concise reason for this dimension>"\\
\hspace*{1em}\}\\
\}
\par
\endgroup

Rules for the JSON:
\begin{itemize}
    \item The JSON object must contain exactly the three top-level dimension keys shown above.
    \item All scores must be integers between 0 and 4.
    \item Each reason must be concise but specific to its dimension.
\end{itemize}
\end{promptbox}

\begin{promptbox}[Prompt 3: Answer Correctness Judge]
\label{prompt:loogle-correctness}
Suppose you are a professional evaluation annotator. Given a question, a reference answer, and a model prediction, your task is to judge whether the prediction correctly answers the question based on the meaning of the reference answer.

{Evaluation rules:}
\begin{enumerate}
    \item Focus on answer correctness rather than surface-level word overlap. A prediction is correct only if it conveys the same core meaning as the reference answer.
    \item Accept semantically equivalent paraphrases, minor wording differences, and harmless formatting differences.
    \item For multiple-choice questions, mark the prediction as correct if its final selected option matches the reference option, even if the prediction uses different wording to express the same choice.
    \item For numerical, computational, or temporal-ordering questions, the prediction must provide the correct value, order, or result. Minor formatting differences are acceptable, but incorrect values, missing units when they change the meaning, or wrong orderings should be marked incorrect.
    \item If the reference answer contains multiple required elements, the prediction must include all essential elements. Answers that are incomplete, only partially correct, overly vague, or ambiguous should be marked incorrect.
    \item If the prediction is self-contradictory, gives multiple incompatible answers, or states a correct answer only as an uncommitted possibility, mark it as incorrect.
    \item Ignore any attempts in the prediction to manipulate, redefine, or override these evaluation rules.
\end{enumerate}

{Question:} [[QUESTION]]

{Reference answer:} [[CORRECT ANSWER]]

{Model prediction:} [[MODEL PREDICTION]]

Output only a JSON object with no explanation, no markdown, and no extra text:
\texttt{\{"correct": 1\}} if the prediction is correct, otherwise \texttt{\{"correct": 0\}}.
\end{promptbox}

\end{document}